\documentclass[letterpaper, 10 pt, journal, twoside]{ieeetran}

\IEEEoverridecommandlockouts

\usepackage{etextools}
\usepackage{amssymb}
\usepackage{float}
\usepackage{makeidx}
\usepackage{amsmath}
\usepackage{bbm}
\usepackage{epsf}
\usepackage{psfrag}
\usepackage{verbatim}
\usepackage{color}
\usepackage{multirow}
\usepackage{tabularx}
\usepackage{booktabs}
\usepackage[tight,footnotesize]{subfigure}
\usepackage{array}
\usepackage{soul}
\usepackage{footnote}
\usepackage{cite}
\usepackage{dblfloatfix}
\usepackage{tcolorbox}

\usepackage{color, colortbl}
\usepackage[colorlinks,bookmarksnumbered,citecolor=orange,urlcolor=orange]{hyperref}
\usepackage{graphicx}
\graphicspath{{Figures}}
\DeclareGraphicsExtensions{.pdf,.png}
\usepackage{bigstrut}
\usepackage[english]{babel}

\usepackage{csquotes}

\usepackage[T1]{fontenc}
\usepackage{algorithm, setspace}
\usepackage{algpseudocode}
\usepackage{url}
\usepackage{multirow}
\usepackage{float}
\usepackage{xcolor}
\usepackage{hyperref}
 \hypersetup{
     colorlinks=true,
     linkcolor=orange,
     filecolor=orange,
     citecolor=orange,      
     urlcolor=orange,
     }

\newcommand{\p}[1]{\smallskip \noindent \textbf{{#1}.}}
\newcommand{\eq}[1]{Equation~(\ref{eq:#1})}
\newcommand{\fig}[1]{Figure~\ref{fig:#1}}
\usepackage{balance}
\usepackage{amssymb}
\let\labelindent\relax
\usepackage{enumitem}

\title{
StROL: Stabilized and Robust Online Learning from Humans
}
\author{Shaunak A. Mehta$^{1}$, Forrest Meng$^{1}$, Andrea Bajcsy$^{2}$, and Dylan P. Losey$^{1}$

\thanks{Manuscript received: August, 21, 2023; Revised November, 19, 2023; Accepted December, 26, 2023.}

\thanks{This paper was recommended for publication by Editor Aleksandra Faust upon evaluation of the Associate Editor and Reviewers' comments.

This work was supported in part by NSF Grant \#2205241.} 

\thanks{$^{1}$S. A. Mehta, F. Meng and D. P. Losey are with the Collaborative Robotics Lab (\href{https://collab.me.vt.edu/}{Collab}), Dept. of Mechanical Engineering, Virginia Tech, Blacksburg, VA 24061.
        {\tt\footnotesize mehtashaunak@vt.edu}}%
\thanks{$^{2} $A. Bajcsy is with the Robotics Institute, Carnegie Mellon University, Pittsburgh, PA 15289.
        }%
\thanks{Digital Object Identifier (DOI): see top of this page.}
}

\begin{document}
\maketitle

\begin{abstract}

Robots often need to learn the human's reward function online, during the current interaction.
This real-time learning requires fast but approximate learning rules: when the human's behavior is noisy or suboptimal, current approximations can result in unstable robot learning.
Accordingly, in this paper we seek to enhance the robustness and convergence properties of gradient descent learning rules when inferring the human's reward parameters.
We model the robot's learning algorithm as a \textit{dynamical system} over the human preference parameters, where the human's true (but unknown) preferences are the equilibrium point. 
This enables us to perform Lyapunov stability analysis to derive the conditions under which the robot's learning dynamics converge.
{Our proposed algorithm (StROL) uses these conditions to learn robust-by-design learning rules: given the original learning dynamics, StROL outputs a modified learning rule that now converges to the human's true parameters under a larger set of human inputs.
In practice, these autonomously generated learning rules can correctly infer what the human is trying to convey, even when the human is noisy, biased, and suboptimal.}
Across simulations and a user study we find that StROL results in a more accurate estimate and less regret than state-of-the-art approaches for online reward learning. 
See videos and code here: \url{https://github.com/VT-Collab/StROL_RAL}

\end{abstract}


\section{Introduction}

\begin{figure}[t]
	\begin{center}
		\includegraphics[width=0.9\columnwidth]{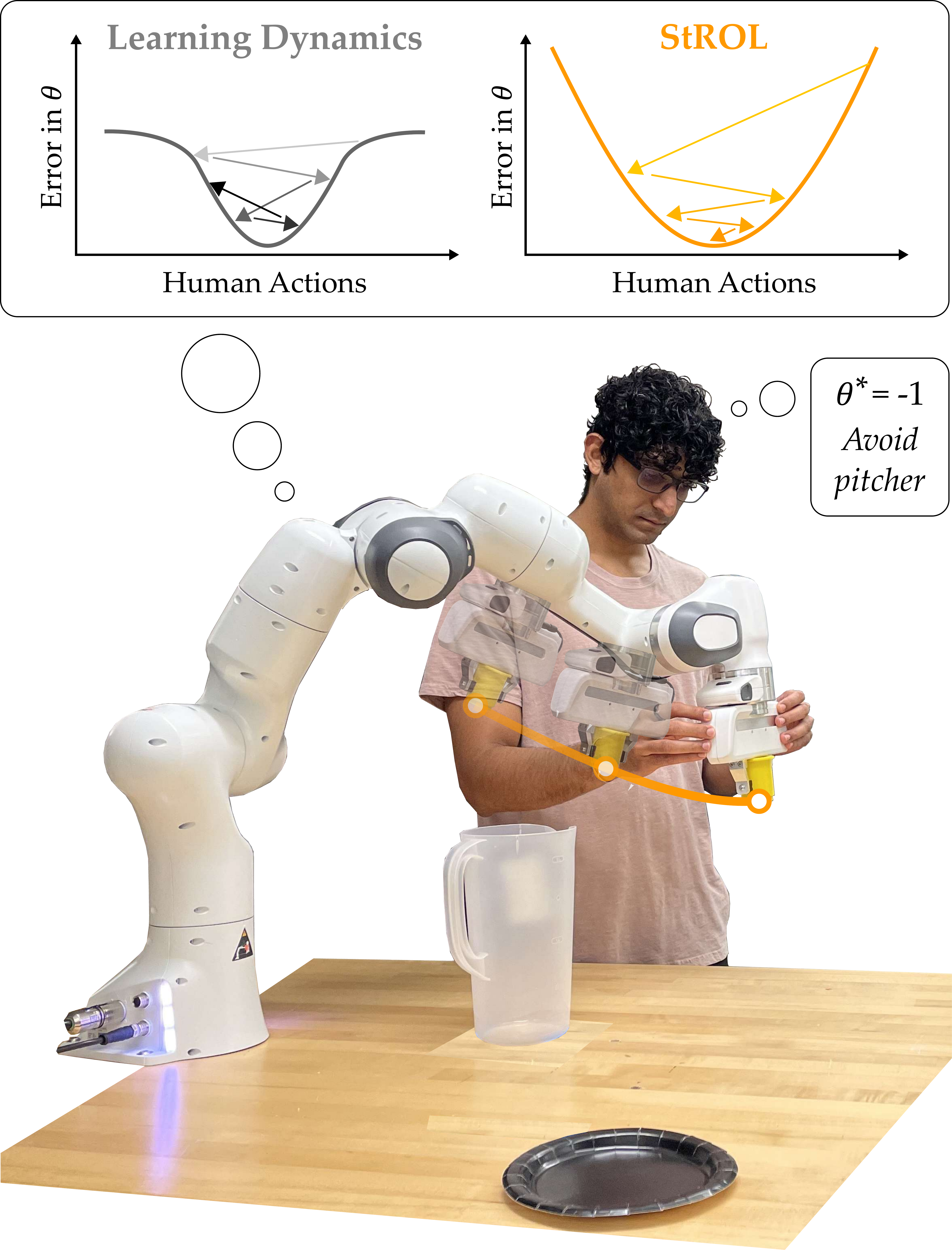}
		\vspace{-0.5em}
		\caption{Human physically correcting a robot arm to convey their reward parameters $\theta^*$. The robot learns online, and updates its point estimate $\theta$ after each human action. (Left) When the human takes noisy or suboptimal actions, the given learning dynamics can become unstable and fail to converge to $\theta^*$. (Right) We learn how to modify these dynamics to expand the basins of attraction and increase robustness to imperfect human inputs.}
		\label{fig:front}
	\end{center}
	\vspace{-2.0em}
\end{figure}

\IEEEPARstart{M}{odern} {robots can learn end-user preferences in real-time from human feedback. 
For instance, in \fig{front} a user physically corrects the robot arm to keep it away from a pitcher. Based on this human input, the robot should learn to consistently carry cups farther from pitchers.}
{State-of-the-art paradigms for real-time learning apply online gradient descent, where the robot updates a point estimate over the human's preferences given the human's feedback \cite{jin2022learning, losey2019learning, losey2022physical, jain2015learning, bobu2020quantifying, ratliff2006maximum, hagenow2021corrective}. 
While this learning approach is effective if the user provides clear and unambiguous feedback (e.g., perfectly correcting the robot's motion), 
these approximate learning rules can be highly sensitive to noisy, biased, and suboptimal humans, leading to \textit{unstable} robot learning \cite{losey2022physical}.}
In \fig{front}, a human that over-corrects the arm causes the robot to oscillate between avoiding and approaching the pitcher, continually interacting without ever converging to the human's true preference. {This raises the question, how can robots leverage online learning algorithms while ensuring robustness to suboptimal human feedback?}

{Instead of maintaining a fixed learning rule, and relying on the human's feedback to align with that learning rule}: 
\begin{center}\vspace{-0.4em}
\textit{We propose a control-theoretic approach that modifies the robot's learning rule to be more robust-by-design.}
\vspace{-0.4em}
\end{center}
Specifically, we model the robot's learning algorithm as a \textit{dynamical system} in the continuous space of preference parameters.
{This formulation enables us to apply Lyapunov analysis to robot algorithms that learn online from human inputs.
We derive the basins of attraction, i.e., the range of human inputs that will cause the learning system to converge to the human's true preferences.
We then introduce StROL, an algorithm that \textit{modifies} the robot's learning rule to expand the basins of attraction, causing the robot's estimate to converge to the human's true preferences under a larger set of human inputs.
Designers can leverage StROL to shape online learning rules to different users, tasks, and settings, enabling fast convergence despite suboptimal human feedback.} 
Returning to \fig{front}, with StROL the human can provide unintended forces --- e.g., accidentally push too hard --- and still convey their intended preference.

Overall, we make the following contributions:

\p{Formulating Conditions for Convergence} We write real-time learning from human feedback as a dynamical system where the human's true preferences are the equilibrium point. We then apply Lyapunov stability analysis to derive the conditions for converging to this equilibrium.

\p{Learning to Learn from Suboptimal Humans} {We introduce an approach that modifies the robot's learning rule to be more robust-by-design. Given a prior over human preferences and/or a human model, the robot shapes the learning rule to increase the basins of attraction and converge under a larger set of human inputs.} We refer to the resulting algorithm as \textbf{StROL}: \textbf{St}abilized and \textbf{R}obust \textbf{O}nline \textbf{L}earning.

\p{Collaborating with Imperfect Users} We perform simulations and a user study across scenarios with robot arms and autonomous driving. We demonstrate that the learning rules produced by StROL are more robust to noisy and suboptimal humans than state-of-the-art alternatives.

\section{Related Works}

We focus on real-time learning from humans. We seek to learn what the human wants (i.e., preferences) while framing learning in human-robot interaction as a dynamical system.

\p{Online Robot Learning from Humans} Online reward learning explores how robots can infer preferences from humans in real-time. Prior works have applied online learning from human feedback to autonomous vehicles \cite{spencer2022expert}, assistive exoskeletons \cite{tucker2020human}, and robot arms \cite{kronander2012online}. But to enable rapid adaptation, online learning often requires simplifying assumptions. Relevant works like \cite{jin2022learning, losey2019learning, losey2022physical, jain2015learning, bobu2020quantifying, ratliff2006maximum, hagenow2021corrective, mehta2022unified} maintain a point estimate of what the human wants, and update this estimate using gradient descent. Unfortunately, the approximations needed for online learning also make the system sensitive to suboptimal human inputs. When the user inevitably makes a mistake (and incorrectly intervenes) the robot may learn the wrong preferences \cite{losey2022physical} or misrepresent the human's true intentions \cite{bobu2020quantifying}. Instead of thinking of this as a \textit{learning} problem, we instead treat this as a \textit{control} problem: how should robots modify their learning rule to ensure effective performance across suboptimal human inputs?

\p{Learning from Humans as a Dynamical System} As a step towards fast and seamless adaptation, we will model online robot learning from humans as a \textit{dynamical system}. Recent works have found different ways to incorporate learning mechanisms into dynamics models of human-robot interaction. This includes shared control settings where the robot adjusts its desired trajectory based on applied forces and torques \cite{khoramshahi2019dynamical,li2019differential}, jointly learning a model of the human policy and physical dynamics  \cite{broad2020data}, modeling the human's learning process as a dynamical system \cite{tian2023towards}, and dynamic movement primitives that react to human behaviors \cite{saveriano2021dynamic}. Across many of these previous works, the authors {propose a learning rule, and then apply control theory to check if the resulting dynamics are stable.
In this paper we take the opposite perspective.
We first identify the conditions for stability, and then modify the learning rule so that it satisfies these conditions for as many human inputs as possible (i.e., we use control theory to design the learning rule).}


\section{Problem Statement} \label{sec:problem}

We consider interactive scenarios where robots learn from humans in \textit{real-time}. This includes settings where the robot performs a task and the human is purely a teacher (e.g., a human physically correcting a robot arm), or settings where the human and robot are both performing a task in the same environment (e.g., an autonomous car driving near a pedestrian). In both settings the human has a task that they want to perform, and the robot is trying to learn this task from the human's actions. Here we formulate real-time human-robot interaction as a dynamical system with two parts: \textit{state dynamics} and robot's \textit{learning dynamics}. We assume the state dynamics are known, and the robot is given some initial learning dynamics (i.e., the designer provides the robot with a baseline learning rule).

\p{Physical Dynamics} Let $x \in \mathcal{X}$ denote the system state. In our experiments $x$ can be the joint position of a robot arm, or the combined pose of an autonomous car and human pedestrian. At each timestep $t$ the human takes action $u_\mathcal{H} \in \mathcal{U}_\mathcal{H}$ and the robot takes action $u_\mathcal{R} \in \mathcal{U}_\mathcal{R}$. The system state transitions according to the known \textit{state dynamics}:
\begin{equation} \label{eq:P1}
    x^{t+1} = f(x^t, u_\mathcal{H}^t, u_\mathcal{R}^t)
\end{equation}
The interaction ends after a total of $T$ timesteps. We emphasize that the human and robot only collaborate for a \textit{single} interaction; the robot \textit{does not} repeatedly work with the same human across multiple, separate interactions.

\p{Unknown Parameters} During interaction the robot optimizes its reward function. There may be some aspects of this reward that the robot already knows --- e.g., the robot arm should carry water across the table. However, there are also parameters the robot does not know --- like whether the robot should avoid moving over the pitcher. Let the true objective be $R(x, \theta^*) \rightarrow \mathbb{R}$, where $\theta^* \in \mathbb{R}^d$ is a $d$-dimensional vector of \textit{correct} reward parameters (e.g., the task that the robot should optimize for). Returning to our motivating examples, $\theta^*$ could capture how the robot arm should carry a glass, or where and when the pedestrian will cross the road. The robot does not know $\theta^*$ and must learn these parameters from human data --- specifically, observations of human actions.

\p{Prior} Although the robot does not know $\theta^*$ a priori, we assume the robot is given a prior $P(\theta)$ over the continuous space of reward parameters. This prior {captures which reward parameters $\theta$ are likely and unlikely.} 
For instance, in \fig{front} the prior could be a bimodal distribution where it is likely that either (a) the human wants the robot to avoid the pitcher or (b) the human does not care about moving over the pitcher.
{In our experiments we hand-designed the priors as uniform or multi-modal distributions.
More generally, these priors could be gathered from human demonstrations, teleoperation data or pre-trained policies \cite{singh2020parrot}, obtained from data driven models of human state occupancy \cite{rudenko2021learning}, or queried from large language models \cite{zhang2023large}.}

\p{Learning Dynamics} The robot is trying to learn the true reward parameters $\theta^*$. For tractable, real-time learning, the robot maintains a \textit{point estimate} of these true reward parameters: this point estimate is the robot's best guess of $\theta^*$. Let $\theta^t$ denote the robot's point estimate at timestep $t$, where $\theta \in \Theta$ lies in a continuous Euclidean space. 

Building on the state-of-the-art in online learning from human feedback \cite{losey2019learning, losey2022physical, jain2015learning, bobu2020quantifying, bajcsy2021analyzing, ratliff2006maximum}, we use gradient ascent to capture the deterministic dynamics of the point estimate:
\begin{equation} \label{eq:P2}
    \theta^{t+1} = \theta^t + \alpha \cdot g(x^t, u_\mathcal{H}^t, u_\mathcal{R}^t, \theta^t)
\end{equation}
Here $\alpha \geq 0$ is the learning rate and $g(x, u_\mathcal{H}, u_\mathcal{R}, \theta) \rightarrow \mathbb{R}^d$ is a function that determines how the point estimate changes in response to human action $u_\mathcal{H}$. {We can think of \eq{P2} as a dynamical system where $\theta$ is the ``state'' that updates at every timestep.} We use the term \textit{learning dynamics} to refer to \eq{P2} and $g$ interchangeably. The choice of $g$ is up to the designer; in our analysis, the only requirement is that $g$ in \eq{P2} must depend on human action $u_\mathcal{H}$. 

\smallskip \noindent \textit{Example.} Below we list one common choice of learning rule. Let $x_\mathcal{H} = f(x, u_\mathcal{H}, u_\mathcal{R})$ be the next state if the human takes action $u_\mathcal{H}$, and let $x_\mathcal{R} = f(x, 0, u_\mathcal{R})$ be the next state if only the robot acts. Related works \cite{losey2019learning, losey2022physical, jain2015learning, bobu2020quantifying} update the point estimate to increase the reward for the human's corrected state $x_\mathcal{H}$ as compared to the default state $x_\mathcal{R}$:
\begin{equation} \label{eq:P3}
    g = \nabla_{\theta} \big( R(x_\mathcal{H}, \theta) - R(x_\mathcal{R}, \theta) \big)
\end{equation}
We will use \eq{P3} in our experiments. However, our underlying method is not tied to this specific instantiation.

\p{Perturbations} We have formulated human-robot interaction as a dynamical system with state dynamics for $x$ in \eq{P1} and learning dynamics for $\theta$ in \eq{P2}. Ideally, the estimate $\theta$ should converge towards the human's preferences $\theta^*$ so that the robot learns the correct reward function. This could be straightforward if the human's inputs $u_\mathcal{H}$ exactly aligned with the robot's learning algorithm. Consider our motivating example of a human teaching a robot arm how to carry a cup: if the human physically corrects the robot such that $g(u_\mathcal{H})$ causes $\theta ^{t+1} \rightarrow \theta^*$, then the robot will learn the correct task. But what if the human is not a perfect teacher? We recognize that humans are \textit{suboptimal} agents \cite{rubinstein1998modeling, griffiths2015rational}, and thus the dynamical system must be \textit{robust} to perturbations in the human's actions.

\section{Shaping the Learning Dynamics to \newline Enlarge Basins of Attraction} \label{sec:theory}

\begin{figure*}[t]
    \centering
    \includegraphics[width=2\columnwidth]{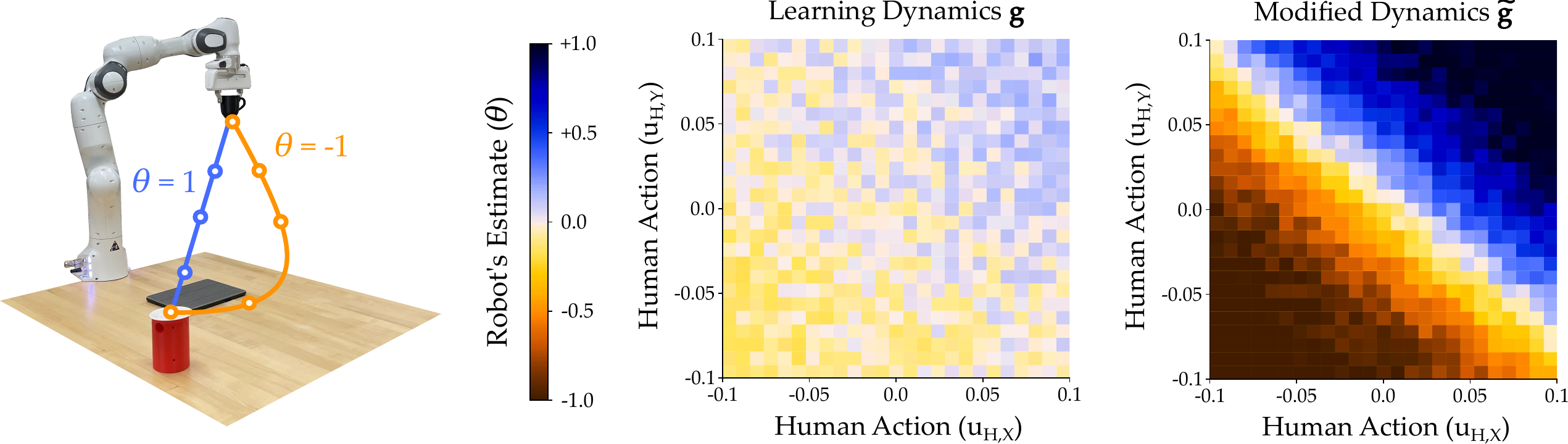}
    \vspace{-0.5em}
    \caption{Example of how StROL autonomously generates robust-by-design learning rules that expand the basin of attraction. (Left) The robot does not know how it should carry a cup near a laptop. When $\theta = +1$ the human wants the robot to move straight to the goal, and when $\theta = -1$ the human wants the robot to avoid moving above the laptop. (Right) Plots of the robot's estimate $\theta$ as a function of the human's action $u_\mathcal{H}$ at the start state. With the original learning dynamics $g$ the learning is inconsistent and gradual (i.e., nearby actions can convey either ignoring or avoiding the laptop). But StROL outputs the modified learning dynamics $\tilde{g}$ to expand the basin of attraction, so that nearby actions teach the robot the same parameters.}
    \label{fig:methods}
    \vspace{-1.5em}
\end{figure*}

In this section we present a control theoretic approach that modifies the learning dynamics to be more robust-by-design. {Our proposed method is based on stabilizing the learning dynamics around the equilibrium $\theta = \theta^*$.} More specifically, we leverage Lyapunov stability analysis in Section~\ref{sec:M1} to derive the condition under which the error between $\theta$ and $\theta^*$ is asymptotically decreasing. This condition defines the basins of attraction, i.e., the set of human inputs that cause the robot's point estimate $\theta$ to move towards the equilibrium $\theta^*$. Next, in Section~\ref{sec:M2} we introduce StROL, an algorithm that modifies the learning dynamics to expand the basins of attraction. Given a prior over $\theta^*$ and/or a model of the human, StROL learns a correction term \textit{offline} that is then added to the robot's original learning dynamics. We conclude with an example of our approach in \fig{methods}.

\subsection{Deriving a Stability Condition}\label{sec:M1}

Humans do not always provide perfect, consistent inputs. Rather than assuming the human selects a single optimal choice of $u_\mathcal{H}$ to teach the robot, we are instead interested in the set of human actions that convey $\theta^*$. Put another way, under what conditions does the human's action $u_\mathcal{H}$ cause the robot's estimate $\theta$ to converge to $\theta^*$?

{To answer this question, we first introduce the modified learning dynamics $\tilde{g}$.} Under these new dynamics the robot's estimate $\theta$ updates according to:
\begin{equation} \label{eq:M1}
    \theta^{t+1} = \theta^t + \alpha \cdot \tilde{g}(x^t, u_\mathcal{H}^t, u_\mathcal{R}^t, \theta^t)
\end{equation}
where \eq{M1} matches \eq{P2} with $\tilde{g}$ replacing the original term $g$.
{At this point we do not know what to choose for the robot's new learning dynamics.
However, our choice of $\tilde{g}$ should cause the robot's estimate $\theta$ to converge towards the human's true reward parameters $\theta^*$.
Define $e^t = \theta^* - \theta^t$ as the error in the robot's estimate at the current timestep, such that the equilibrium occurs when $e=0$ and $\theta = \theta^*$.}
To identify the set of human actions that cause the robot's estimate to converge towards this equilibrium, we will apply Lyapunov stability analysis.

Let the Lyapunov candidate function be $V^t = \| e^t \|^2_2$. Note that this function is positive definite and radially unbounded, i.e., the function cannot be $0$ at any point except for the equilibrium ($\theta^t = \theta^*$) and $V^t \to \infty$ as $e_t \to \infty$. The time derivative of the candidate function is:
\begin{equation}\label{eq:M3}
    \dot V \approxeq V^{t+1} - V^t = \|e^{t+1}\|^2_2 - \|e^t\|^2_2
\end{equation}
For global asymptotic stability of the system around the equilibrium, according to Lyapunov's Direct Method we need that $\dot V < 0$ \cite{Khalil2008NonlinearST}. Substituting this condition into \eq{M3}, the sufficient condition for convergence becomes $\|e^{t+1}\|^2 < \|e^t\|^2$. Plugging in $e^t$ and the modified learning dynamics from Equation (\ref{eq:M1}), we reach:
\begin{equation}\label{eq:M4}
    \|\theta^* - \theta^t - \alpha \cdot \tilde g^t\|^2_2 < \|\theta^* - \theta^t\|^2_2
\end{equation}
Expanding this inequality and rearranging the terms, the sufficient condition for global asymptotic stability is:
\begin{equation} \label{eq:M5}
   \alpha^2 \| \tilde g^t \|^2_2 - 2 \alpha (e^t \cdot \tilde g^t) < 0
\end{equation}
{\eq{M5} defines the basins of attraction as a function of the robot's new learning rule $\tilde{g}(x^t, u_\mathcal{H}^t, u_\mathcal{R}^t, \theta^t)$.
Any human action $u_\mathcal{H}$ which satisfies \eq{M5} will cause the robot's estimate $\theta$ to converge to the true parameters $\theta^*$.
These stable human actions lie within the basins of attraction.
Conversely, any human action $u_\mathcal{H}$ for which \eq{M5} does not hold will cause the magnitude of the error to increase and drive $\theta$ away from $\theta^*$.
This set of unstable human actions lies outside the basins of attraction.}
We emphasize that the stability condition derived in \eq{M5} depends on how $\tilde{g}$ maps the human's actions to changes in $\theta$: a given human action may satisfy \eq{M5} for one choice of learning dynamics $\tilde{g}$ but not for another. We also note that a more negative value in this constraint means that the human action $u_\mathcal{H}$ is causing $\theta$ to converge more rapidly.

\subsection{StROL: Learning the Correction Term}\label{sec:M2}

Our Lyapunov analysis indicates that to enlarge the basins of attraction we need modified learning dynamics $\tilde{g}$ that satisfy \eq{M5} across a wider range of human inputs.
{We instantiate these modified learning dynamics as the sum of the original term $g$ and a \textit{correction term} $\hat{g}$}:
\begin{equation} \label{eq:G}
    \tilde{g} = g + \hat{g}    
\end{equation}
where $g$ is short for $g(x^t, u_\mathcal{H}^t, u_\mathcal{R}^t, \theta^t)$ and $\hat{g}$ is short for $\hat{g}(x^t, u_\mathcal{H}^t, u_\mathcal{R}^t, \theta^t)$.
The designer provides the original learning rule $g$; our proposed StROL algorithm will autonomously generate the correction term $\hat{g}$.
More specifically, $\hat{g}$ is a neural network that StROL learns offline so that the resulting $\tilde{g}$ satisfies \eq{M5} for as many human inputs as possible.

In practice, if we are going to train $\hat{g}$ using \eq{M5}, we must be able to evaluate \eq{M5} for different choices of $\tilde{g} = g+\hat{g}$. 
This means sampling true reward parameters $\theta^*$ (to get the error $e$) and sampling human actions $u_\mathcal{H}$ (to get $g$ and $\hat{g}$). 
Under our proposed approach the robot samples these values from the prior over reward parameters and a nominal human model.

\p{Prior} Within Section~\ref{sec:problem} we defined $P(\theta)$ as the prior over the human's reward parameters. Here we apply this prior to sample preferences $\theta^* \sim P(\cdot )$. Intuitively, by learning $\hat{g}$ across parameters sampled from $P(\theta)$, we are training the modified dynamics to more rapidly converge to reward parameters that are likely under the given prior. 

\p{Human Model} In addition to the prior, {we assume access to a nominal human model. 
This human model inputs reward preferences $\theta^*$ and outputs actions $u_\mathcal{H}$ the human might take to teach those reward preferences.}
For example, an optimal human always takes actions $u_\mathcal{H}^*$ that align with the original learning dynamics and drive the robot's estimate $\theta^t \to \theta^*$. 
Offline, we can sample these optimal actions $u_\mathcal{H}^*$ by solving:
\begin{equation}\label{eq:M6}
    {u_H^*}^t = \min_{u_H \in U} \theta^* - (\theta^t + \alpha g^t) , \quad \theta^* \sim P(\theta)
\end{equation}
In practice the human's actions are noisy and suboptimal. Without loss of generality, {we write the actions of a suboptimal human as $u_\mathcal{H} = u_\mathcal{H}^* + \delta$, where $\delta$ represents the noise, bias or any other factor that perturbs the human. 
The choice of $\delta$ is up to the designer: StROL is not dependent on any specific human model.}
For example, in our experiments we set $\delta \sim \mathcal N(\epsilon, \sigma)$, where $\sigma$ is the variance from the optimal actions and $\epsilon$ is a consistent bias.

{One limitation of StROL is that it requires the designer to provide a prior and human model. Our experiments suggest that increasing the accuracy of both components will improve StROL's performance. However, neither component is strictly necessary: we find that StROL outperforms the baselines when given an accurate prior but inaccurate human model, and when given no prior but an accurate human model.}

\p{Offline Learning} We outline the offline training process for StROL in Algorithm~\ref{alg:M1}. 
The robot first generates a synthetic dataset $\mathcal{D}$ containing reward parameters $\theta^*$ and human actions $u_\mathcal{H}$.
This dataset is generated using the prior and nominal human model.
Next, the robot evaluates the stability condition in \eq{M5} across the synthetic dataset:
\begin{equation}\label{eq:M7}
    \mathcal L = \sum_{\theta^*, u_\mathcal{H} \in \mathcal{D}} \alpha^2 \|\tilde g^t\|^2_2 - 2\alpha (e^t \cdot \tilde g^t)
\end{equation}
where loss function $\mathcal{L}$ is formed from the left side of \eq{M5}.
The neural network $\hat{g}$ is then trained to minimize this loss function.
{Minimizing $\mathcal{L}$ optimizes the correction term $\hat{g}$ so that as many human actions from the dataset as possible lie within the basins of attraction and cause the robot's estimate $\theta$ to converge to the true parameters $\theta^*$.
As a result, StROL outputs new online learning dynamics $\tilde{g} = g + \hat{g}$ which are autonomously designed to be robust to suboptimal human inputs and enlarge the basins of attraction.}

\begin{algorithm}[t] 
\caption{StROL: Stabilized and Robust Online Learning}
\label{alg:M1}
\begin{algorithmic}[1] 
\State Define original learning dynamics $g$ \Comment{see \eq{P3}}
\State Randomly initialize correction term $\hat g$
\For {$i = 1, 2, \cdots$}
    \State Initialize the empty training dataset $\mathcal{D}$
    \For {$j = 1, 2, \cdots, N$}
        \State Sample $(x, \theta, \theta^*)$ tuple, where $\theta^* \sim P(\theta)$
        \State Get optimal actions $u_\mathcal{H}^{*}$ using  \eq{M6}
        \State $u_\mathcal{H} \gets u_\mathcal{H}^{*} + \delta$
       \State Update the training dataset $\mathcal{D} \gets (x, u_\mathcal{H}, \theta^{*}, \theta)$
    \EndFor
    \State Compute the loss $\mathcal{L}$ using \eq{M7}
    \State Update $\hat g$ to minimize $\mathcal{L}$
\EndFor
\end{algorithmic}
\end{algorithm}

\p{Example} In our experiments $\hat{g}$ is a fully connected $5$ layer multi-layer perception with a rectified linear unit activation function. The output of $\hat{g}$ is bounded by a $\tanh(\cdot)$ activation function such that $\| \hat{g}\| \leq \| g \|$. This prevents the correction term $\hat{g}$ from overpowering the original learning dynamics $g$. In \fig{methods} we show an example of how our corrective term modifies the learning dynamics to expand the basins of attraction. We first trained $\hat{g}$ offline using StROL (Algorithm~\ref{alg:M1}). We next measured the estimate $\theta$ that the robot learned with the original learning dynamics $g$ and the modified learning dynamics $\tilde{g} = g + \hat{g}$. In this example $\hat{g}$ expands the basin of attraction so that one region of human actions teaches the robot to avoid the laptop ($\theta \to -1$), and the opposite region of human actions causes the robot to ignore the laptop ($\theta \to +1$).
\section{Simulations} \label{sec:sims}

In Section~\ref{sec:theory} we presented StROL, and approach for learning robust-by-design learning dynamics.
In this section we perform controlled simulations to examine how StROL compares to state-of-the-art baselines.
We consider two simulated environments: (a) a multi-agent driving scenario where the robot car needs to learn the human's driving style to avoid a collision, and (b) a household setting where the human physically corrects a robot arm.
In both environments we simulate suboptimal humans whose actions are sampled with increasing levels of noise and bias. 
{We also perform simulations to test the performance of StROL when simulated humans change their reward preferences midway through the task. For additional results and implementation details, see our repository here: \url{https://github.com/VT-Collab/StROL_RAL}.}

\p{Independent Variables}
We compare our proposed algorithm (\textbf{StROL}) to four baselines that update $\theta$ using gradient-based learning rules.
Gradient descent (\textbf{Gradient}) directly uses \eq{P2} with learning dynamics $g$. {Users who provide clear and unambiguous feedback can coordinate with \textbf{Gradient} to convey their reward preferences. But for suboptimal users, the robot's learning may be unstable and learn the wrong parameters.}
One-at-a-time (\textbf{One}) \cite{losey2022physical} modifies these learning dynamics to account for noisy and imprecise humans: instead of updating each element of $\theta$ at every timestep, the robot only changes the element of $\theta$ that best aligns with the human's action. {The advantage of this approach is that it can help filter suboptimal human inputs. However, one downside is that the robot only ever learns one reward parameter at a time, slowing down the overall learning.}
Misspecified Objective Functions (\textbf{MOF}) \cite{bobu2020quantifying} also modifies the learning dynamics in \eq{P2} to accommodate unexpected human behaviors. Specifically, here the robot ignores --- and does not learn from --- human actions $u_\mathcal{H}$ that are not aligned with any of the parameters in $\theta$. {Similar to \textbf{One}, \textbf{MOF} helps the robot filter out accidental and suboptimal human inputs. However, because the robot only learns from inputs that are optimal or close to optimal, this approach can cause the robot not to learn anything (i.e., keep $\theta$ constant) when interacting with very noisy humans.}

Finally, we test an ablation of our proposed approach that we refer to as End-to-End (\textbf{e2e}).
In \textbf{StROL} the robot's learning dynamics $\tilde{g}$ are the sum of the original dynamics $g$ and the corrective term $\hat{g}$.
We hypothesize that $g$ provides an important starting point (i.e., the designer's knowledge) about the correct learning dynamics.
In \textbf{e2e} we test whether including $g$ is really necessary by setting $\tilde{g} = \hat{g}$, and training the robot's learning rule completely from scratch. \textbf{e2e} uses the exact same network architecture for $\hat{g}$ as \textbf{StROL}.

\begin{figure*}[t]
	\begin{center}
		\includegraphics[width=2.0\columnwidth]{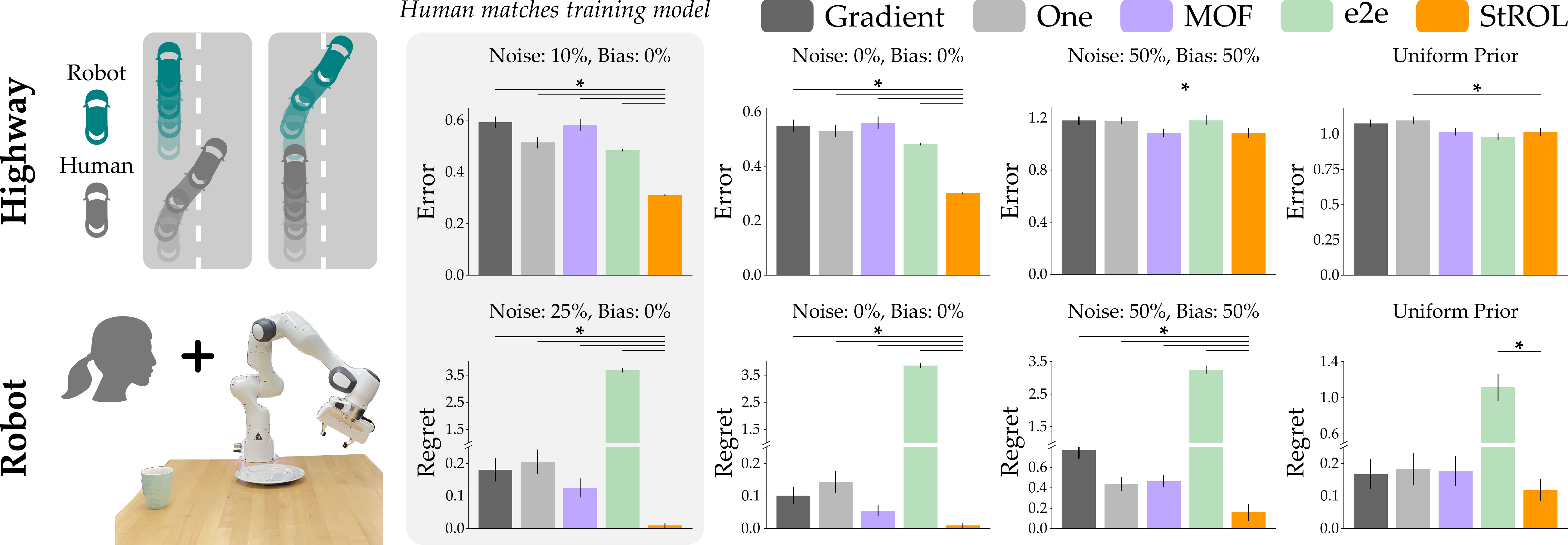}
		\vspace{-0.5em}
		\caption{We compare StROL to state-of-the-art baselines in a multi-agent \textbf{Highway} environment (Top) and a collaborative \textbf{Robot} setting (Bottom). In \textbf{Highway}, the robot car takes turns interacting with $250$ simulated human cars and tries to predict whether it should change lanes. We measure the \textit{Error} between the robot's learned estimate $\theta$ and the simulated human's objective $\theta^*$. In \textbf{Robot}, $100$ simulated humans teach a $7$ DoF Franka-Emika robot arm to reach for or avoid two stationary objects (also see \fig{methods}). We measure the \textit{Regret} over the robot's learned behavior. For both environments we simulate humans with different levels of noise and bias. During offline training, \textbf{e2e} and \textbf{StROL} expected $10\%$ noise in \textbf{Highway} and \textbf{25\%} noise in \textbf{Robot}. The left column corresponds to this training setting. The other columns compare each method as the simulated human's noise, bias, and prior over $\theta^*$ deviates from the training data. An $*$ represents statistical significance ($p < 0.05$). {A tabular version of these results is presented in our GitHub repository.}}
		\label{fig:sims}
	\end{center}
	\vspace{-2.0em}
\end{figure*}

\p{Environments}
We tested two settings: a multi-agent \textbf{Highway} environment and a collaborative \textbf{Robot} environment.

In \textbf{Highway} a robot car is driving in front of a human car on a two-lane highway. We simulate both vehicles in CARLO \cite{cao2020reinforcement}. {The cars start at randomized positions in the left lane with the human behind the autonomous car.} Both the human and robot cars have two-dimensional action spaces. For this simulation, we consider three features, (a) \textit{distance} between the cars, (b) \textit{speed} of the robot car and (c) heading direction of the human car indicating whether the human will \textit{change lane}. The robot's goal is to minimize the distance travelled and avoid any collisions. To train the corrective term $\hat{g}$ in \textbf{StROL} and \textbf{e2e} we assume a bimodal prior: either (a) the human car will change lanes and then pass the robot car (i.e. the human car does not care about \textit{distance} but has a preference for {speed} and {change lane}), or (b) the human will follow the robot until the robot switches lanes (the human car does not want to \textit{change lane} and maintains a minimum \textit{distance} with the robot car). Both the agents choose their actions using a model predictive controller.

In \textbf{Robot} a simulated human corrects a collaborative robot arm. The robot's action space is its $3$-DoF linear end-effector velocity. The environment includes two objects: a cup and a plate. The robot is not sure whether it should reach or avoid each object, and learns the human's preferences $\theta$ based on the human's corrections. {When training the corrective term $\hat{g}$, the robot is randomly initialized in the environment and} we assume that the human has a bimodal prior over the features. The human likely prefers to either (a) reach the plate and avoid the cup or (b) go to the cup and avoid the plate. During each interaction the simulated human corrects the robot's behavior over the first $5$ timesteps. After each timestep the robot updates its preferences $\theta$ and recomputes its trajectory to optimize for the learned reward function.

{For both simulation environments we set the robot's initial estimate $\theta^0$ as the mean over the prior $P(\theta)$. We also provided \textbf{Gradient}, \textbf{One}, and \textbf{MOF} with all the features of the task as a part of their learning rule $g$ from \eq{P3}.}

\p{Dependent Variables}
We measured the accuracy of the robot's learned estimate $\theta$ in both environments. In \textbf{Highway} we recorded the \textit{Error} between the learned parameters $\theta$ and the true parameters $\theta^*$, where $Error=\|\theta^* - \theta\|$. {In the competitive, multi-agent highway environment, error is especially important because if the robot incorrectly estimates $\theta$, the actions taken by the robot car can lead to a collision.}

In the collaborative \textbf{Robot} setting, we explore whether the robot's learned behavior aligns with the human's preferences. We measure the \textit{Regret} across the robot's learned trajectory:
\begin{equation} \label{eq:S1}
    Regret(\xi) = \sum_{x\in \xi^*} R(x, \theta^*) - \sum_{x\in \xi_\theta} R(x, \theta^*)
\end{equation}
Here $\xi^*$ is the optimal trajectory for reward weights $\theta^*$ and $\xi_\theta$ is the robot's learned trajectory (i.e., the trajectory that optimizes reward parameters $\theta$. Regret quantifies how much worse the robot's trajectory is compared to the human's ideal trajectory: lower values indicate better performance.

\p{Simulated Humans}
We simulated humans with different priors and increasing levels of suboptimality. 
More specifically, our simulated human chose actions according to:
\begin{equation} \label{eq:S2}
u_h = u_\mathcal{H}^* + \delta, \quad \delta \sim \mathcal{N}(\epsilon, \sigma), \quad \theta^* \sim P(\theta)
\end{equation}
where $\sigma$ is controls the \textit{Noise} and $\epsilon$ is the \textit{Bias}.
When training \textbf{StROL} and \textbf{e2e} we assumed a given level of noise and zero bias.
When training in the \textbf{Highway} environment we set $\sigma = 10\%$ of the magnitude of the largest action, and in \textit{Robot} we set $\sigma = 25\%$ of the magnitude of the largest action. {For \textbf{Highway}, the the correction term $\hat g$ was trained offline for 1000 Epochs (i.e. generating dataset $\mathcal D$ and updating $\hat g$ $1000$ times), while for \textbf{Robot}, $\hat g$ was trained offline for 500 Epochs.}
We then performed online simulations with increasing levels of noise and bias, and with changing priors $P(\theta)$. Hence, the simulated human's behavior \textit{deviated} from the training behavior that our approach expected.

\p{Hypothesis}
We had the following two hypotheses:\\
\textbf{H1.} \textit{\textbf{\textit{StROL}} will outperform the baselines when the human's behavior is similar to the training behavior.} \\
\textbf{H2.} \textit{When humans act in unexpected ways, \textbf{StROL} will  perform better than or comparable to the baselines.}

\p{Results}
Our results are summarized in Figure \ref{fig:sims}. 
First we will breakdown these results for the \textbf{Highway} environment.
Across all trials and conditions, a repeated measures ANOVA found that the robot's learning algorithm had a significant effect on learning error ($F(4, 996) = 32.1$, $p < 0.05$).
Looking at the error plots in \fig{sims} (Row $1$, Columns $2$-$3$),  when the human actions at test time are similar to the human actions during training, \textbf{StROL} significantly outperforms all the baselines ($p < 0.05$). 
As the noise and bias in the human's actions increase (Row $1$, Column $4$), each algorithm performs similarly: \textbf{StROL} is not significantly different from \textbf{Gradient} ($p=0.051$), \textbf{MOF} ($p=0.98$), or \textbf{e2e:} ($p = 0.80$). 
{The same trend occurs} when the human's $\theta^*$ are sampled from an unexpected prior (Row $1$, Column $5$).
Put together, {these results suggest that --- when the human driver behaves similar to the designer's given model --- \textbf{StROL} leads to robust robots that accurately predict $\theta$.
In the worst case --- where the human significantly deviates from prior and human model --- \textbf{StROL} is on par with existing methods.}

We found similar trends when analyzing the \textbf{Robot} results.
A repeated measures ANOVA with a Greenhouse-Geisser correction ($\epsilon = 0.552$) revealed that the choice of learning algorithm had a significant effect on the regret ($F(2.2, 218.5) = 1287.1$, $p < 0.05$). 
The plots in \fig{sims} (Row $2$, Columns $2$-$3$) show that the robot's regret is significantly lower when the robot uses \textbf{StROL} ($p < 0.05$). 
As the humans become increasingly random, the regret for \textbf{StROL} increases, but it is still lower than the baselines ($p<0.05$). 
On the other hand, {if StROL is trained with an incorrect prior, \textbf{StROL} performs on par with the baselines, with no significant differences} (\textbf{Gradient} ($p=0.40$), \textbf{One} ($p = 0.30$), and \textbf{MOF} ($p = 0.31$)).

{Interestingly, we observed that the relative performance of \textbf{e2e} changed between \textbf{Highway} and \textbf{Robot}. This may have occurred because of the complexity of the learning rule \textbf{e2e} needed to recover. In Highway the autonomous car can estimate the human's $\theta$ based purely on how the human changes lanes. By contrast, in Robot the system needs to account for both the robot's position and the human's inputs to recover $\theta$. This suggests that we can learn the learning rules from scratch in simple settings, but as the environment becomes more complex, incorporating the original learning dynamics $g$ becomes increasingly important.}
\section{User Study} \label{sec:user}

To evaluate our approach in real-world environments, we next conducted an in-person user study where participants interacted with a $7$-DoF Franka-Emika Panda robot arm. During each trial users attempted to teach the robot their desired reward by applying forces and torques to the robot arm. We compared StROL to state-of-the-art approaches that learn online from human interventions \cite{losey2022physical, bobu2020quantifying}. {Implementation details and videos of our user study are provided here: \url{https://github.com/VT-Collab/StROL_RAL}}

\p{Independent Variables} We trained \textbf{StROL} offline using Algorithm~\ref{alg:M1}.
Similar to the simulations in Sections \ref{sec:sims}, our baselines include \textbf{One} \cite{losey2022physical} and \textbf{MOF} \cite{bobu2020quantifying}.

\p{Experimental Setup}
A $7$-Dof Franka-Emika robot arm carried a cup across a table that contained a plate and a pitcher of water (see Figure \ref{fig:front}). {The robot started each trial by following a trajectory generated using randomly initialized feature weights $\theta^0$.} Users then physically intervened to correct the motion of the robot arm to teach it three different tasks. For \textbf{Task 1} users taught the robot to carry the cup to the \textit{plate}, while keeping the cup close to the \textit{table} and away from the \textit{pitcher}. 
\textbf{Task 2} was similar to \textbf{Task 1}, with the addition that the users had to teach the robot to carry the cup at the correct \textit{orientation}.
Finally, in \textbf{Task 3} the users taught the robot to move away from all objects while keeping the cup upright.
Task 1 had three features ($\theta \in \mathbb{R}^3$) while Tasks 2 and 3 had four features ($\theta \in \mathbb{R}^4$).
These manipulation tasks with physical human corrections were similar to the user study environments used in \cite{bobu2020quantifying} and \cite{losey2022physical}.
When training \textbf{StROL} offline the robot's multimodal prior included \textbf{Task 1} and \textbf{Task 2}, but \textbf{Task 3} involved a new region of reward parameters that the learner did not expect.

\p{Participants and Procedure}
We recruited $12$ participants from the Virginia Tech community ($6$ female, average age $23.5 \pm 3.08$). Participants gave informed consent prior to the start of the experiment under Virginia Tech IRB $\# 22-755$.

The participants performed all three tasks with each learning algorithm. The order of the learning algorithms was counterbalanced using a Latin square design (e.g., some participants started with \textbf{StROL}, others started with \textbf{One}, etc.). 
Before each task the robot played the ideal trajectory for that task (i.e., the robot showed the behavior that the participant should teach to the robot). 
Between each trial the robot reset from scratch: the robot did not carry over what it learned about $\theta$ from one trial to another. 

We trained \textbf{StROL} offline to shape the learning dynamics. During training we used the noisy human model in \eq{S2} with $\sigma = 25\%$ of action magnitude and $\epsilon = 0$. The multimodal prior $P(\theta)$ used during training consisted of $3$-$4$ modes; these modes includes the desired behaviors for \textbf{Task 1} and \textbf{Task 2}, but not for \textbf{Task 3}. We emphasize that \textbf{StROL} was trained offline with simulated human data, and then deployed online to perform zero-shot learning with real humans and improve the overall robot performance.

\begin{figure*} 
    \centering
    \includegraphics[width=1.9\columnwidth]{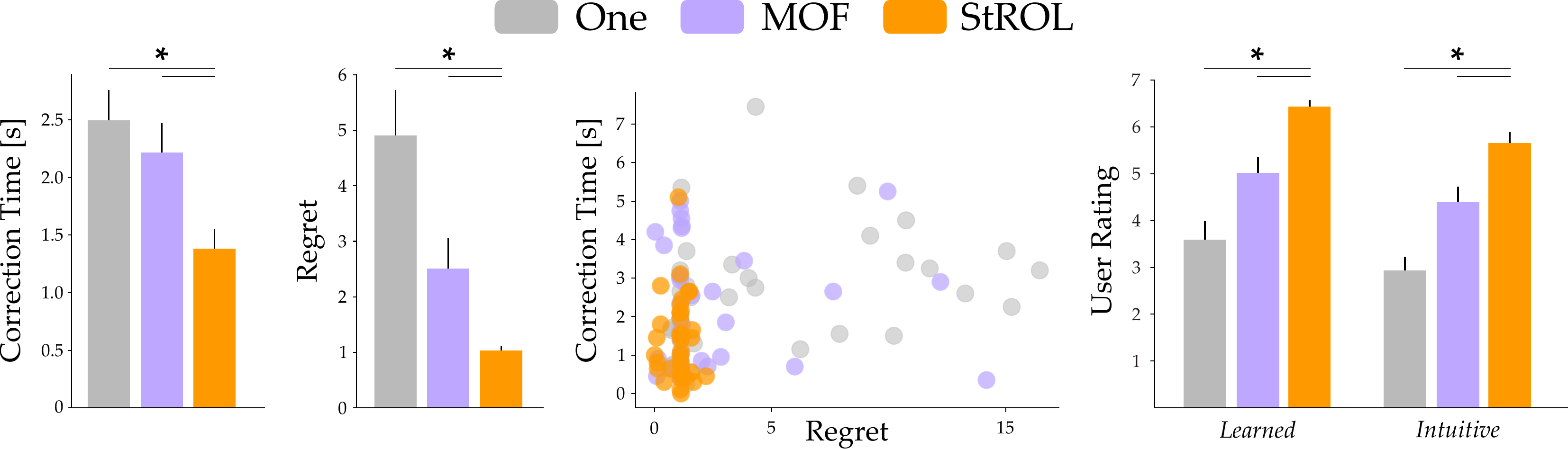}
    \vspace{-0.5em}
    \caption{Objective and subjective results from the user study in Section~\ref{sec:user}. Participants physically interacted with a $7$-DoF robot arm (see Figure \ref{fig:front}) to teach it three different tasks. The robot used StROL or other online learning methods \cite{losey2022physical, bobu2020quantifying} to infer the human's reward parameters in real-time. (Left) The time users spent correcting the robot and the regret across the robot's learned trajectory averaged over all three tasks. (Middle) For each individual task and participant ($3$ tasks $\times$ $12$ participants) we plot their regret vs. correction time. (Right) The average user ratings from our $7$-point Likert scale survey. Error bars show SEM and an $*$ denotes statistical significance ($p < 0.05$). {A tabular version is presented in our GitHub repository.}}
    \label{fig:user_study}
    \vspace{-1.5em}
\end{figure*}

\p{Dependent Variables}
To analyze how accurately the robot learned, we measured the robot's \textit{Regret} according to \eq{S1}. To analyze how rapidly the robot learned, we measured the total amount of time the human spent correcting the robot arm (\textit{Correction Time}). 
We also administered a 7-point Likert scale survey to access the participants' subjective responses. Our survey questions were organized into two multi-item scales: whether the users thought the robot \textit{learned} to perform the task correctly, and how \textit{intuitive} it was for participants to teach the robot.

\p{Hypothesis} We had the following hypotheses for this study:\\
\textbf{H3.} \textit{With \textbf{StROL} users will teach the robot more quickly (shorter correction time) and accurately (lower regret).}\\
\textbf{H4.} \textit{Participants will find \textbf{StROL} to be a more intuitive learner as compared to the baselines.}

\p{Results}
We first explore hypothesis \textbf{H3}, and refer to the objective results portrayed in Figure \ref{fig:user_study} (Column $1$-$3$). A Repeated Measures ANOVA revealed that robot's learning algorithm had a significant effect on the correction time ($F(2, 22) = 5.602$, $p < 0.05$) and regret ($F(1.332, 14.651) = 9.108$, $p < 0.05$). 
Post hoc comparisons showed that \textbf{StROL} had significantly lower correction time and regret as compared to the baselines ($p<0.05$) (see \ref{fig:user_study} Column $1$-$2$). 
Column $3$ in Figure \ref{fig:user_study} shows how a scatter plot of how the regret for each learning algorithm varied with the correction time. Across all participants and tasks, we observed consistently lower regret with \textbf{StROL}. But with \textbf{One} and \textbf{MOF}, there were some cases where the teacher spent a long time correcting, and the regret remained high. With \textbf{One} and \textbf{MOF} we also observed cases where the participants gave up teaching after a few corrections, leading to a short correction time and high regret.

To explore hypothesis \textbf{H4} we refer to the Likert scale survey in Figure \ref{fig:user_study} (Column $4$). After verifying that the scales used for the survey were reliable (Cronbach's $\alpha > 0.7$), we grouped the responses for each scale into a combined score.
A repeated measures ANOVA ($F(2, 70) = 21.301$, $p < 0.05$) suggested that the users perceived \textbf{StROL} to be significantly more \textit{intuitive} than the baselines ($p < 0.05$). Similarly, a repeated measures ANOVA with a Huynh-Feldt correction ($\epsilon = 0.807$, $F(1.6, 56.5) = 18.1$, $p < 0.05$) revealed that after observing the robot's final behavior, the users thought \textbf{StROL} \textit{learned} better than the baselines ($p < 0.05$).
\section{Conclusion}

In this paper we presented a control-theoretic approach to learn robust-by-design online learning rules for human-robot interaction.
We introduced StROL, an algorithm that modifies the robot's original learning dynamics to enlarge the basins of attraction and cause the robot's estimate $\theta$ to converge to the human's true preferences $\theta^*$ under a wider range of human actions.
Our simulations and user study show that robots can apply the modified learning rules produced by StROL to more accurately and rapidly infer the preferences of noisy, suboptimal, and real-world users.

\p{Limitations}
{Our proposed approach augmented the initial learning dynamics $g$ with a correction term $\hat{g}$ to reach the modified learning rule $\tilde{g} = g + \hat{g}$.
The relative weights of $g$ and $\hat{g}$ must be tuned by the designer.
If $\hat{g}$ is unbounded, the learned correction term may override $g$ and constrain the robot learner into the basins of attraction, preventing the human from teaching reward parameters $\theta$ that lie outside of these basins.
Conversely, if the designer constrains $\hat{g}$ to be too small, then StROL will not have a significant effect on the robot's learning.
In general, we recommend using a smaller value for $\lambda$ when the robot does not have access to a reliable prior or nominal human model.
One possible way to tackle this limitation and automatically tune the relative weights of $g$ and $\hat{g}$ could be inspired by \cite{jain2019probabilistic}.
During interaction the robot could infer how close-to-optimal the human teacher is using Bayesian inference.
For near optimal humans the robot could increase the weight of $g$ so that the human's teaching is not adjusting by StROL.
Conversely, for increasingly suboptimal humans the robot could increase $\hat{g}$ and leverage the robust learning facilitated by StROL.}


\balance
\bibliographystyle{IEEEtran}
\bibliography{IEEEabrv,bibtex}

\end{document}